\documentclass[sigconf, nonacm]{acmart}
\usepackage{graphicx} 

\renewcommand\footnotetextcopyrightpermission[1]{}
\acmConference{}{} 
\acmBooktitle{} 

\usepackage[]{hyperref} 
\usepackage{bm}
\usepackage{xspace}
\usepackage{pifont}
\usepackage{color}
\usepackage{amsmath}
\usepackage{makecell}
\usepackage{wrapfig}
\usepackage{booktabs}
\usepackage[skip=3pt]{subcaption}
\usepackage{enumitem}
\usepackage{microtype}
\usepackage[normalem]{ulem} 
\usepackage[linesnumbered, ruled,vlined]{algorithm2e}
\usepackage{trimspaces}
\usepackage{listings}

\usepackage{xcolor}

\definecolor{codegreen}{rgb}{0,0.6,0}
\definecolor{codegray}{rgb}{0.5,0.5,0.5}
\definecolor{codepurple}{rgb}{0.58,0,0.82}
\definecolor{backcolour}{rgb}{0.95,0.95,0.92}

\lstdefinestyle{jsonstyle}{
    backgroundcolor=\color{backcolour},
    basicstyle=\ttfamily\footnotesize,
    breakatwhitespace=false,
    breaklines=true,
    captionpos=b,
    frame=single,
    numbers=left,
    numbersep=5pt,
    showspaces=false,
    showstringspaces=false,
    showtabs=false,
    tabsize=2,
    rulecolor=\color{black}
}

\newcommand{\narrow}[1]{}

\makeatletter
\g@addto@macro{\UrlBreaks}{\UrlOrds}
\makeatother
\makeatletter
\g@addto@macro{\UrlBreaks}{\UrlOrds}
\makeatother

\newcommand{\sysname}{RAGPulse\xspace}
\newcommand{\twen}{Twen\xspace}

\newif\ifshowrevise
\showrevisetrue  

\newcommand{\revise}[2]{%
  \ifshowrevise
    \ifx&#1&\else{\color[RGB]{0,0,192}\sout{#1}}\fi  
    \ifx&#1&\else\ \fi  
    \ifx&#2&\else{\color[RGB]{192,0,0}{#2}}\fi  
  \else#2\fi 
}

\usepackage[textsize=small]{todonotes}

\newif\ifshowtodos
\showtodostrue  

\newcommand\paraspace{\vspace*{0.25ex}}

\providecommand\parae[1]{\paraspace\textbf{\textit{#1}}}

\newcommand{\etc}{\emph{etc.}\xspace}
\newcommand{\ie}{\emph{i.e.,}\xspace}
\newcommand{\eg}{\emph{e.g.,}\xspace}

\newcommand{\secref}[1]{\S\ref{#1}}
\newcommand{\figref}[1]{Figure~\ref{#1}}

\newcommand{\listref}[1]{List~\ref{#1}}

\title{\sysname: An Open-Source RAG Workload Trace to Optimize RAG Serving Systems}

\usepackage{etoolbox}
\makeatletter
\patchcmd{\authornote}{\g@addto@macro\addresses{\@authornotemark}}{}{}{}
\makeatother

\begin{document}









\author{Zhengchao Wang, Yitao Hu, Jianing Ye, Zhuxuan Chang, Jiazheng Yu, Youpeng Deng, Keqiu Li}

\authornote{Corresponding author: Yitao Hu (email: yitao@tju.edu.cn).}
\affiliation{%
  \institution{Tianjin University, China}
  \country{}
}
\begin{abstract}
Retrieval-Augmented Generation (RAG) is a critical paradigm for building reliable, knowledge-intensive Large Language Model (LLM) applications. However, the multi-stage pipeline (retrieve, generate) and unique workload characteristics (\eg knowledge dependency) of RAG systems pose significant challenges for serving performance optimization. Existing generic LLM inference traces fail to capture these RAG-specific dynamics, creating a significant performance gap between academic research and real-world deployment.

To bridge this gap, this paper introduces \sysname, an open-source RAG workload trace dataset. This dataset was collected from an university-wide Q\&A system serving that has served more than 40,000 students and faculties since April 2024. We detail \sysname's system architecture, its privacy-preserving hash-based data format, and provide an in-depth statistical analysis. Our analysis reveals that real-world RAG workloads exhibit significant temporal locality and a highly skewed hot document access pattern. \sysname provides a high-fidelity foundation for researchers to develop and validate novel optimization strategies for RAG systems, such as content-aware batching and retrieval caching, ultimately enhancing the efficiency and reliability of RAG services. The code is available at \href{https://github.com/flashserve/RAGPulse}{https://github.com/flashserve/RAGPulse}.
\end{abstract}

\maketitle
\section{Introduction}

Large Language Models (LLMs)~\cite{osdi24llumnix, Fast25_Mooncake, asplos24_SpecInfer} have demonstrated powerful capabilities across a spectrum of natural language processing tasks. However, their practical application is often hindered by two inherent limitations: \textit{knowledge cutoff}, where LLM models are unaware of information post-dating their training, and \textit{hallucination}, where LLM models confidently generate incorrect, nonsensical, or fabricated information.

\begin{figure}[t]
    \centering
    \includegraphics[width=1.0\linewidth]{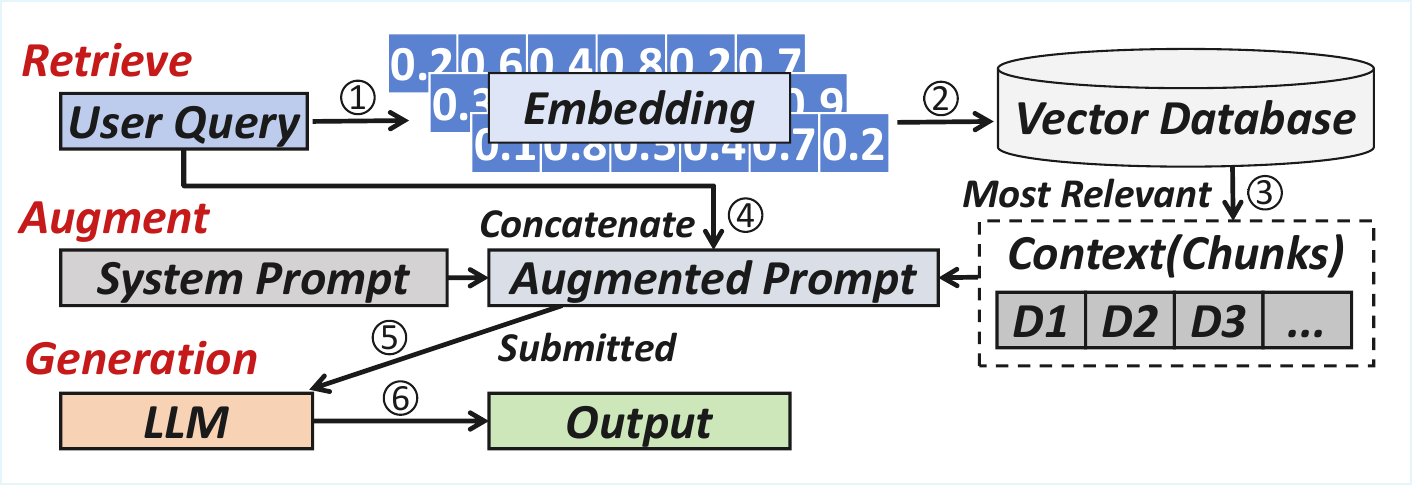}
    \caption{Workflow of RAG.}\narrow{}
    \label{fig:rag_overview}
\end{figure}

Retrieval-Augmented Generation (RAG)~\cite{sosp24_ragserve, NIPS24_RankRAG, ICLR25_Turborag} is an advanced paradigm designed to address these challenges. The core idea of RAG is to synergize the powerful reasoning and generative capabilities of LLMs with the real-time, factual accuracy of external knowledge bases. As shown in \figref{fig:rag_overview}, instead of directly tasking LLM with an answer, the RAG workflow employs a multi-stage pipeline:

\begin{itemize}
    \item Retrieve: Once the RAG service system receives a request, an embedding model vectorizes the user query and retrieves the most semantically relevant knowledge chunks from a large-scale vector database (\eg FAISS~\cite{dataset24faiss}).
    \item Augment: These retrieved chunks are then concatenated as "context" with the original user query and a system prompt.
    \item Generate: Finally, this "augmented prompt" is submitted to the LLM, guiding it to produce a factually-grounded and accurate output based on the provided real-time context.
\end{itemize}


By decoupling knowledge storage from the model's reasoning capabilities, the RAG paradigm significantly enhances the reliability, timeliness, and traceability of LLM applications. It has rapidly become the standard for building enterprise-grade Q\&A systems, intelligent document analysis tools, and trustworthy AI assistants (such as the university-wide policy Q\&A service in this study).

However, despite its significance, optimizing the performance and cost of RAG serving systems presents a new set of challenges. Unlike pure LLM inference, the performance bottlenecks in RAG systems are composite. Performance is not only dependent on the generation stage (\eg KV cache efficiency) but is also heavily influenced by the retrieval stage (\eg vector database latency) and the complex interactions between these stages (\eg adaptive batching).

We find that existing generic LLM inference traces fail to capture the unique workload dynamics specific to RAG~\cite{BurstGPT, ShareGPT-Chinese-English-90k}. As our subsequent analysis reveals (\secref{sec:analysis}), real-world RAG workloads exhibit highly skewed knowledge chunk access frequencies and significant inter-request temporal locality (\ie a \textit{hot document} phenomenon). These characteristics are critical for designing efficient retrieval caching and content-aware batching strategies, yet they are entirely absent from generic LLM traces.

This performance gap between research tools (generic traces) and practical deployments (RAG systems) severely hinders optimization research for RAG serving. Academia and industry urgently require a public benchmark dataset that accurately reflects real-world RAG workload characteristics, which is the primary focus of this paper.
\section{Preliminary and Motivation}

\subsection{Why We Need a RAG-Specific Trace?}

Existing generic LLM inference traces exhibits significant limitations when applied to RAG serving system. This inadequacy stems from the unique characteristics of RAG workloads, which differ fundamentally from pure LLM inference and necessitate the development of RAG-specific traces.

First, RAG possesses inherent multi-stage pipeline complexity. A RAG request must sequentially flow through distinct phases, including retrieval, re-ranking, generation, and so on~\cite{NIPS24_RankRAG, iclr24_SelfRAG, ICLR25_Turborag}. This architecture dictates that system latency, resource consumption, and caching behaviors are governed by complex inter-stage interactions. Generic LLM traces, which typically focus only on the generation phase, fail to characterize these composite effects.

Second, RAG performance demonstrates strong dependencies on the knowledge base and query patterns. Its workload dynamics, such as query similarity, embedding cache hit rates, and retrieval latency, fluctuate significantly with real-world user behavior. Furthermore, in many practical applications, conseutive queries often exhibit significant inter-request contextual dependencies. Capturing these dynamics and inter-request correlations is critical for optimizing KV cache reuse, batching, and scheduling strategies~\cite{Agarwal2025CacheCraftMC, arxiv24_Ragcache}.

Moreover, RAG-specific traces are fundamental to enabling the co-optimization of retrieval and generation components. Traditional approaches that treat retrieval and generation as independent black boxes restrict systemic design potential. In contrast, RAG trace data can inform more sophisticated system designs, such as dynamic retrieval caching or query-adaptive batching~\cite{wang-etal-2024-rag}.

Finally, existing RAG trace data, often derived from synthetic or genetic chat workloads, creates a performance gap between RAG optimization research and practical deployment. Evaluating optimization strategies using non-representative traces leads to a disconnect between research findings and real-world system performance. This discrepancy poses severe challenges to the efficiency, stability, and reliability of RAG serving systems.

Therefore, a real-world RAG trace is needed to advance the practical application of RAG research.

\subsection{What We Could Do with a RAG Trace?}
\begin{figure}[t]
    \centering
    \includegraphics[width=1\linewidth]{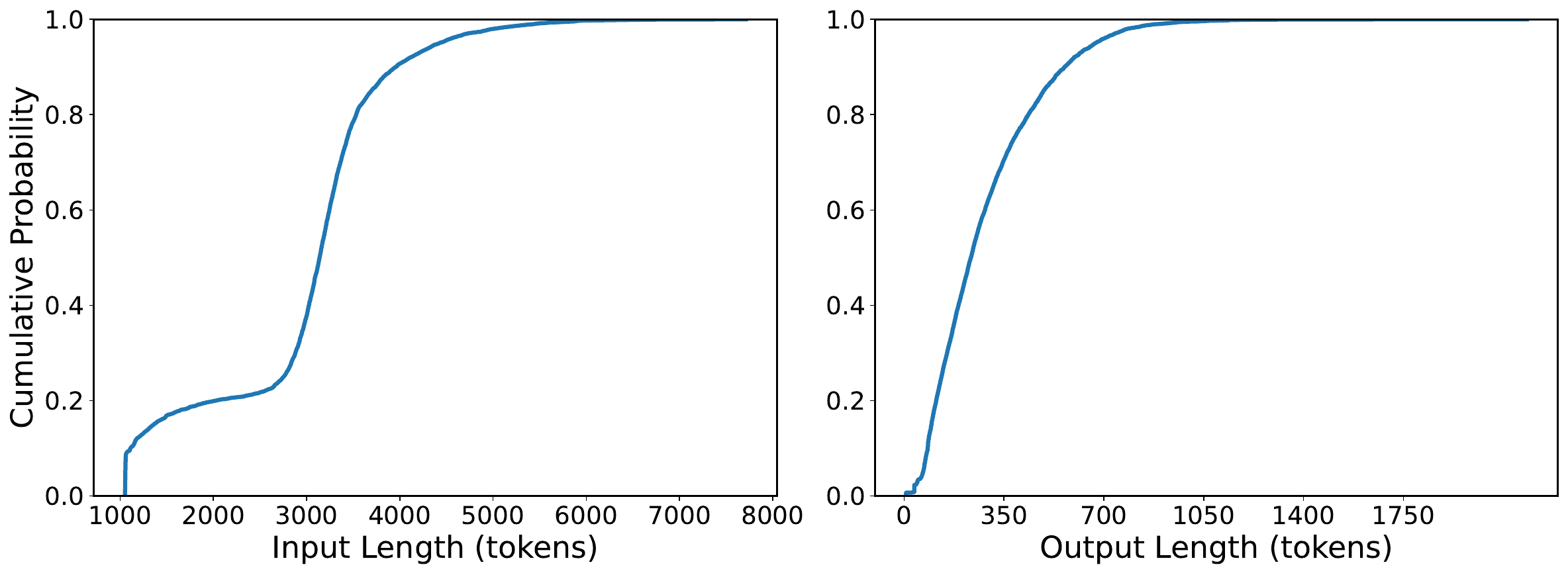}
    \caption{CDF of Input and Output Token Lengths in RAGPulse.}\narrow{}
    \label{fig:cdf}
\end{figure}

A RAG-specific trace dataset that reflects realistic workload characteristics would be a critical tool for research and optimization. It would unlock several key pathways for investigation and application in both academia and industry:

\parae{Precise Performance Bottleneck Analysis.} The trace would allow researchers to conduct fine-grained breakdowns of end-to-end latency in RAG systems, precisely quantifying the respective latency contributions and bottlenecks of the retrieval, reranking, and generation stages.

\parae{Informed Optimization of Scheduling and Caching Policies.} It would provide a realistic foundation for evaluating and designing advanced system strategies, including KV cache reuse efficiency, adaptive batching algorithms, and retrieval caching, all under practical inter-request correlation patterns.
    
\parae{High-Fidelity Workload Modeling and Benchmarking.} The data would provide the foundation for building high-fidelity RAG workload models, simulators, and standardized benchmarks, thereby ensuring the validity and reproducibility of system evaluations.

\parae{Study of Emerging Application Behaviors.} It would support the exploration of advanced RAG applications, such as tracking and analyzing how real-world workloads evolve from simple Q\&A patterns into more complex, agent-like reasoning and multi-turn task-solving behaviors.

These applications show how a high-fidelity RAG trace would serve as a key enabler for bridging the gap between theoretical RAG systems research and practical deployment, driving advancement in the field.

\section{Introduction to RAGPulse}

To bridge the gap between RAG systems research and practical deployment, we introduce RAGPulse, a RAG workload trace dataset collected from a real-world deployment. The dataset originates from an university-wide policy Q\&A system, which has been in continuous operation since April 2024, serving over 40,000 students and faculties. We are now open-sourcing the core of RAGPulse as a real-world RAG trace. We are committed to continuously updating this dataset as our service evolves, with plans to release further RAG and future Agent traces to reflect evolving workload patterns.

\subsection{Data Format and Content}

Each record in the RAGPulse trace captures key system-level runtime information for a RAG request. The dataset we are releasing at this time covers 7,106 request records sampled from one week of the system's operation. As shown in \figref{fig:cdf}, the input and output token lengths of the RAGPulse workload demonstrate a clear concentration trend. Specifically, input token lengths primarily cluster around 3500 tokens, while output token are concentrated around 500 tokens. Furthermore, as detailed in \listref{tab:ragpulse_record}, each record in the main trace file contains the following key fields:

\begin{table}[htbp]
    \centering
    \begin{minipage}{0.95\linewidth}
    \begin{lstlisting}[caption={Request Component Sample.}, label={tab:ragpulse_record}]
{   
    "timestamp": "27", 
    "input_length": 3861, 
    "output_length": 127, 
    "hash_ids": {
        "sys_prompt": [8325, 8326, 11575], 
        "passages_ids": [6123, 7239, 6124, 1167, 7250, 5448], 
        "history": [15215], 
        "web_search": [20319, 20320], 
        "user_input": [23648]
        }, 
    "session_id": "1758081660427-xa8rbsd2uco1"
}
    \end{lstlisting}
    \end{minipage}
\end{table}

\begin{itemize}
    \item \texttt{timestamp}: It denotes the request submission time, which is calculated in seconds starting from the trace's initial moment (12:00:00).
    \item \texttt{input\_length}: The total input token length of the request. This is the sum of the token lengths of all components included in the request (\eg system prompt, retrieved document, chat history).
    \item \texttt{output\_length}: The total token length of the model-generated output.
    \item \texttt{hash\_ids}: A comprehensive collection of hash identifiers representing every component of the request's input. This includes unique IDs for the the system prompt, all retrieved documents, user chat history, external web search results, and user's question.
    \item \texttt{session\_id}: The conversation identifier to which the request belongs.
\end{itemize}


To rigorously protect user privacy, all original textual content has been removed and replaced with remapped hash\_ids. This approach ensures complete anonymization while preserving all the structural and temporal characteristics required for research, making it highly suitable for system performance and scheduling policy analysis.


\subsection{Dataset Features and Statistical Analysis}
\label{sec:analysis}
\begin{figure}[t]
    \centering
    \includegraphics[width=1\linewidth]{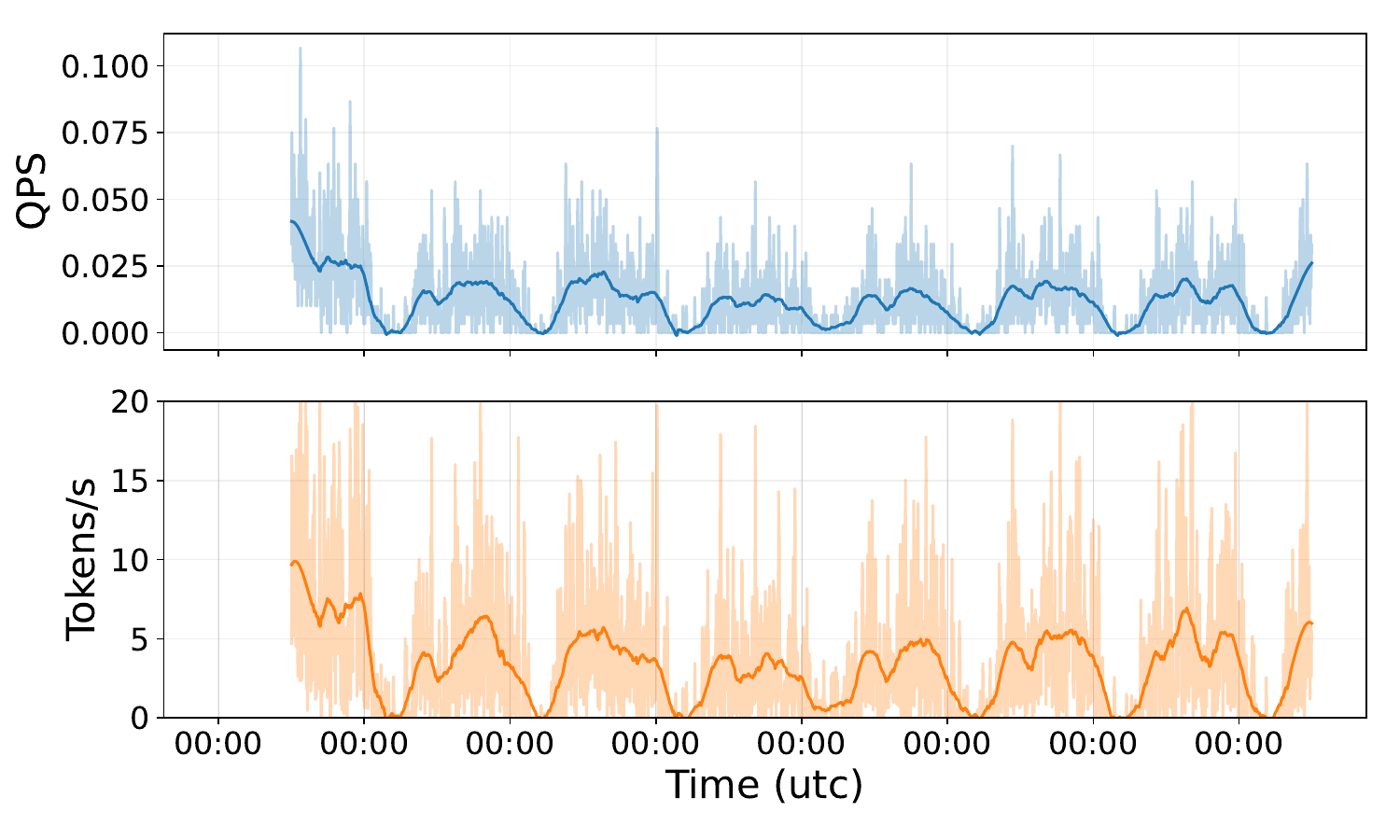}
    \caption{Throughput over time in RAGPulse.}\narrow{}
    \label{fig:Throughput}
\end{figure}
A preliminary statistical analysis of the RAGPulse trace data reveals four key characteristics that are highly instructive for RAG serving system design.

\subsubsection{Periodic Load Fluctuation}
\ 
\newline
\indent RAGPulse records the arrival time of each request. As illustrated in \figref{fig:Throughput}, the workload exhibits clear periodicity. The request volume peaks during daytime (working hours) and decreases significantly at night. This pattern is highly consistent with the real-world user behavior of serving systems~\cite{Fast25_Mooncake, BurstGPT}.

\subsubsection{Dynamic Input Composition}
\begin{figure}[t]
    \centering
    \includegraphics[width=0.9\linewidth]{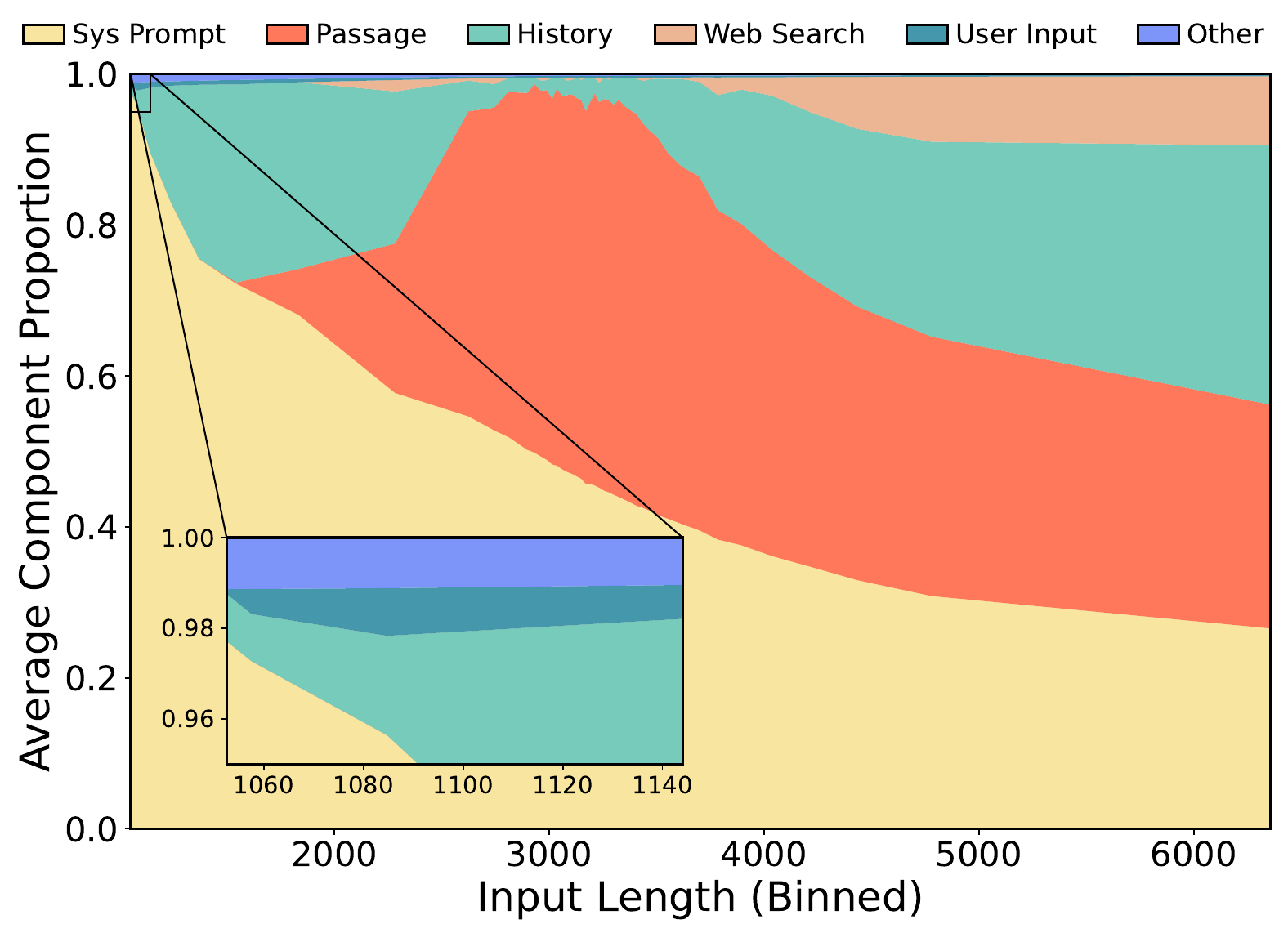}
    \caption{Proportion of Input Components Across Different Input Lengths in RAGPulse.}\narrow{}
    \label{fig:components}
\end{figure}

\ 
\newline
\indent As shown in \figref{fig:components}, we further analyzed the composition of the request input tokens. The analysis reveals that the proportional contribution of different components is not static, but varies dynamically with the total input length.

Specifically, for shorter requests (with a total input length of less than 2000 tokens), the system\_prompt is the dominant component, accounting for over 70\% of the tokens. In these cases, the contribution from retrieved passages and history is minimal. However, as the total input length increases (exceeding 3000 tokens), the load composition shifts significantly. The token contribution from retrieved passages and history (dialogue history) increases substantially, collectively approaching 40\% of the total.

This dynamic input composition provides a critical insight for RAG system optimization: the system's processing overhead (such as the memory and computation for context processing) is not uniform, but is highly heterogeneous and dependent on the request's length and type.

\subsubsection{Text Usage Frequency}
\ 
\newline
\indent When analyzing the usage of the hash\_ids field within the trace, we observed a significant phenomenon: the reuse rate of hash\_ids corresponding to retrieved document from the vector database is far higher than that of other components.

Specifically, as shown in \figref{fig:characteristics}(a), a small subset of core knowledge chunks ("hot" documents) is frequently retrieved and referenced by a large volume of different user requests, while the vast majority of chunks are accessed much less often. This highly skewed usage distribution exhibits a typical "long-tail effect." This finding provides strong evidence that Retrieval Caching, which targets these retrieved content chunks, is a highly valuable optimization direction for RAG systems.


\subsubsection{Document Overlap between Requests}
\begin{figure}[t]
    \centering
    \includegraphics[width=1\linewidth]{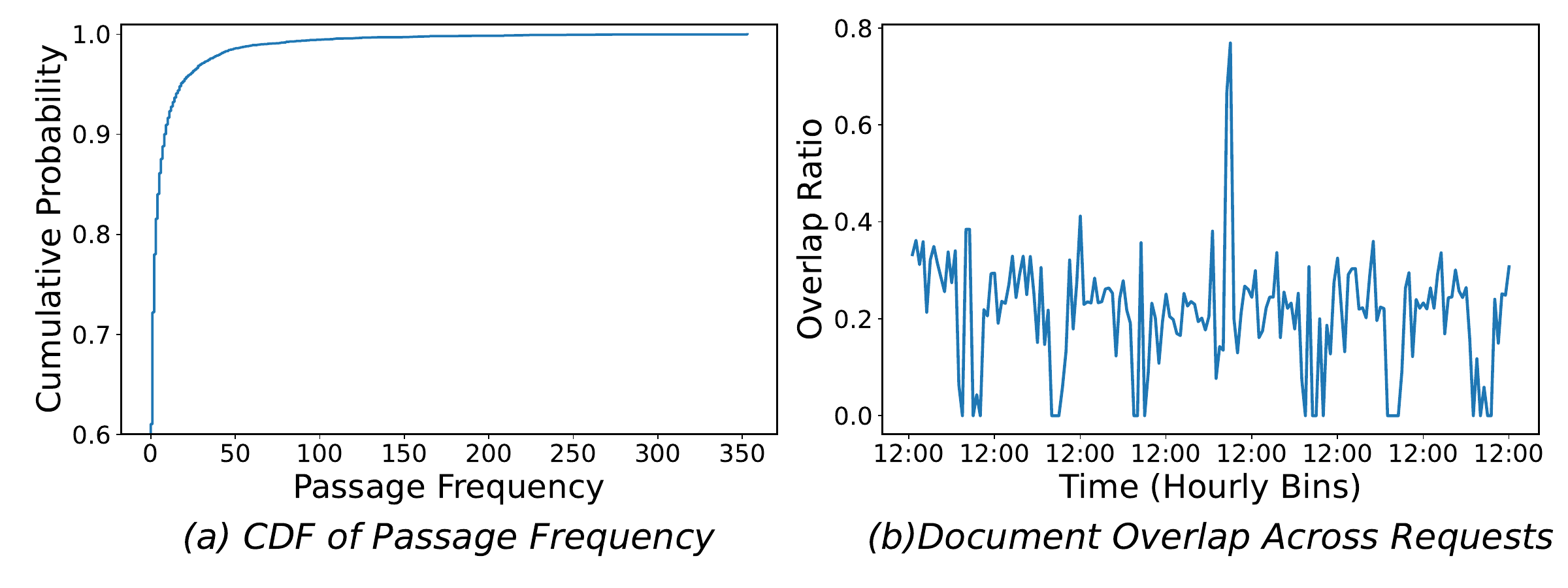}
    \caption{Partially Unique Characteristics of RAG Traces in RAGPulse.}\narrow{}
    \label{fig:characteristics}
\end{figure}

\ 
\newline
\indent Further analysis of the request time-series reveals significant temporal locality. As shown in \figref{fig:characteristics}(b), within any given time window, a large proportion of concurrently arriving requests tend to access the same or highly similar hash\_ids for knowledge chunks.

This phenomenon results in high inter-request document overlap. This confirms our hypothesis regarding strong inter-request contextual dependencies in real-world RAG workloads. This characteristic indicates that the system could gain significant benefits from two strategies: (1) Content-aware batching of requests that access the same documents; and (2) the corresponding KV Cache for these "hot" documents will have exceptionally high reuse potential.


\section{Dataset Usage and Applications}

To validate the effectiveness of our proposed RAGPulse dataset for evaluating system performance in real-world environments, we have constructed a dedicated benchmarking platform. This chapter provides a detailed description of the platform's specific implementation, the composition and usage methodology of the dataset, and the performance evaluation metrics employed.

\subsection{Implementation}
We constructed a prototype system for an online RAG service to simulate real-world applications. The system's core is built upon vLLM~\cite{kwon2023efficient}, a widely adopted open-source inference framework. To emulate a typical RAG workflow, the system integrates an end-to-end pipeline, encompassing user request reception, text processing, and final result generation. All experiments were conducted on a server equipped with a single NVIDIA A800 GPU (80GB VRAM). For the Generator component of the RAG pipeline, we deployed Qwen2.5-14B as the LLM model.

\subsection{Scaling RAGPulse to Any Scale}
RAGPulse's trace data is characterized by a significant temporal span\footnote{Currently, we are releasing one week's data, with plans for longer-period data in the future.}. This property provides researchers with high flexibility, allowing them to slice or scale the trace data according to specific experimental goals. For example, a high-traffic period (\eg one hour) can be isolated to benchmark the system's peak concurrent processing capabilities. Conversely, the entire long-term trace can be used to assess the long-term efficacy of caching policies, such as the KV cache management. Moreover, an analysis of throughput fluctuations in the trace enables a deeper comprehension of the system's behavioral patterns under varying loads.

\subsection{Dataset Text Decryption}
To protect user privacy, all original text within the dataset has been replaced by hash IDs. Nonetheless, for system-level simulations, especially for the evaluation of components reliant on specific text content like KV cache, the characteristics of the original text (such as length) are essential. Therefore, researchers must employ a text reconstruction strategy when using this dataset. The central principle of this strategy is to generate a unique, random-content placeholder text for every unique hash ID, ensuring the token length is strictly identical to that of the original text. It is imperative that the same hash ID maps consistently to the exact same placeholder text in all requests, thereby guaranteeing the simulation's consistency and validity.

\subsection{Metrics and Setups}
To comprehensively evaluate the performance of the RAG serving system, we adopted two key metrics widely recognized in the field of LLM serving:

\begin{itemize}
    \item Time To First Token (TTFT): This measures the total time elapsed from the moment the system receives a request until the first output token is generated and returned. It is the core benchmark for system responsiveness and user-perceived latency.
    \item Time Per Output Token (TPOT): This measures the average time required to generate each subsequent token after the first token has been produced. This metric primarily reflects the system's throughput and processing efficiency during the generation phase.
\end{itemize}

\subsection{Demo Use}
The RAGPulse benchmark utilizes a client-server architecture to implement its workload replay mechanism.

The server-side component, built on the vLLM framework, manages the core inference tasks. It receives client requests via an OpenAI-compatible API and supports both streaming and non-streaming generation modes. This capability is essential for capturing key performance metrics, including TTFT and TPOT. The server employs an asynchronous, concurrent design to maintain stability and high throughput under heavy load.

The client-side component acts as the load generator, driving the entire benchmark execution. Its three-stage workflow begins with Data Preprocessing, where the client reads the trace file and reconstructs text inputs by generating random token sequences that strictly match the original token lengths for each hash ID. Next, the Request Scheduling module dispatches the processed requests to the server, strictly adhering to the original timestamps from the trace file. To enhance flexibility, a time-scaling factor is introduced, allowing researchers to accelerate or decelerate the workload replay. Finally, the Metric Collection module records the per-request performance data, such as TTFT and TPOT, and ensures results are persistently stored for subsequent analysis.


Please refer to \url{https://github.com/flashserve/RAGPulse/blob/main/example/single_online_instance/README.md} for details.

\begin{figure}[t]
    \centering
    \includegraphics[width=0.9\linewidth]{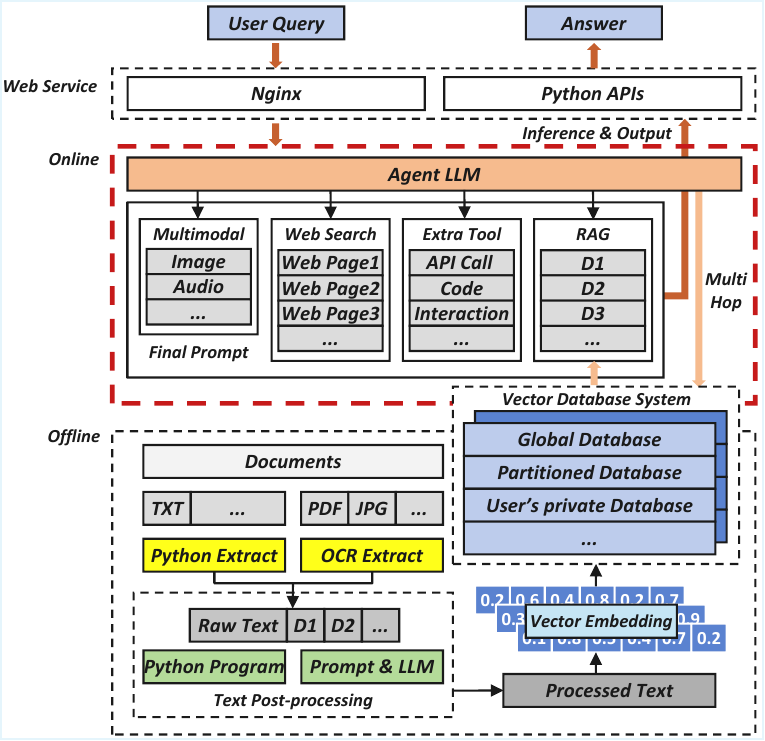}
    \caption{\twen's System Architecture.}\narrow{}
    \label{fig:sys_overview}
\end{figure}

\section{Data Source: \twen System Architecture}

The \sysname trace dataset, which is sampled from \href{https://twen.ai/}{Twen.ai}, an university-wide Q\&A RAG system that has served over 40,000 students and faculties since April 2024. This section details \twen's system architecture to provide context for how the trace data was generated.

\subsection{System Overview}
The core design philosophy of \twen is the separation of retrieval and generation. To achieve this, \twen adopts a Python-based microservice architecture (FastAPI, Docker) to decouple the CPU-intensive retrieval tasks from the GPU-intensive generation tasks. The overall system architecture is illustrated in \figref{fig:sys_overview}:

\begin{itemize}
    \item Generator Model: We use Qwen3-235B-A22B-Instruct-2507 model for main Q\&A and agent interaction functions~\cite{qwen}. 
    \item Embedding Model: We use the locally deployed open-source model infgrad/stella-large-zh-v3-1792d for vectorizing all knowledge base documents.
    \item Orchestration: LangChain is used to build and orchestrate prompt templates, model call chains, and retrieval flows~\cite{anthropic2024langchain}.
    \item Vector Database: Qdrant~\cite{Qdrant} is used to store and index knowledge base vectors with a tag-based structure, organized into global, tagged, and user-personal databases.
    \item Document Maintenance: The system uses Python programs and LLM for content extraction, format standardization, and automatic segmentation of files.
\end{itemize}


\subsection{Offline Stage}

\begin{figure}[t]
    \centering
    \includegraphics[width=0.8\linewidth]{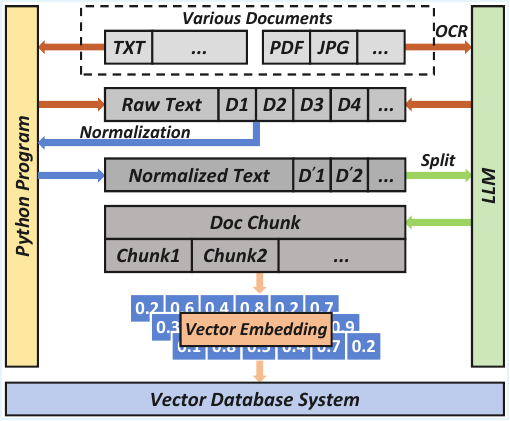}
    \caption{Offline Architecture in \twen.}\narrow{}
    \label{fig:offline}
\end{figure}

As shown in \figref{fig:offline}, the core task of the offline stage is to construct the high-quality vector knowledge base.

The knowledge base corpus (university policies and information) is acquired via three channels: (1) Manual submission by faculty and department leaders; (2) An upload system for authorized administrators; and (3) Targeted web crawlers that periodically pull data from public department websites.

Based on their storage formats, the raw documents are categorized into two types: the first consists of files like TXT, from which content can be easily extracted by Python programs; the second includes files like JPG, where content is not explicitly stored and requires extraction using LLM-based OCR. The extracted raw content often has messy formatting. To address this, we developed an specialized agent for text normalization. The normalized text is then intelligently segmented using another dedicated LLM\footnote{We use a variety of LLM models ranging from 7B to 70B for different sub-tasks for cost efficiency.} to ensure semantic integrity, resulting in high-quality corpus chunks suitable for building the vector database.

\subsection{Online Stage}

\begin{figure}[t]
    \centering
    \includegraphics[width=0.6\linewidth]{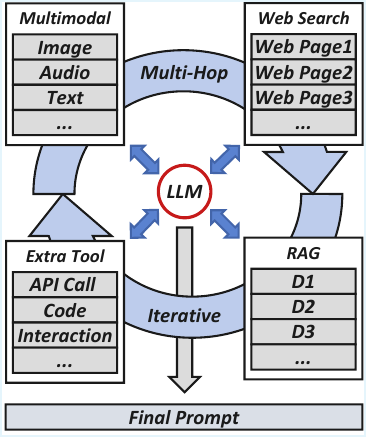}
    \caption{Online Architecture in \twen.}\narrow{}
    \label{fig:online}
\end{figure}

As shown in \figref{fig:online}, the online stage handles user requests and generates the \sysname trace records in the process.

We have developed our agent system, a comprehensive framework where a central large language model, which we refer to as "the Agent LLM," acts as an autonomous controller. When a request arrives, this Agent LLM autonomously plans the optimal processing strategy based on user information and prompts. It can interpret queries from multimodal perspectives (images, audio, text, \etc) and dynamically orchestrates a suite of tools—including web search, RAG, and a series of predefined external tools (API calls, programming, \etc) to comprehensively solve problems. Furthermore, the system architecture supports proactive self-reflection: the Agent LLM first assesses its own response, and if it deems the current answer insufficient, it will iteratively employ various tools in a multi-hop manner until reaching the final response.


The system calls the Qwen3-235B model asynchronously in streaming mode using LangChain's \texttt{AsyncIteratorCallbackHandler}. During this call, the Langfuse \texttt{CallbackHandler} is activated, capturing the complete call metadata (including latency, token lengths, user\_id, session\_id, and retrieved hash\_ids). This captured data is sent to the Langfuse database, forming the \sysname trace dataset analyzed in this study.

\section{Conclusion}
RAG has emerged as the standard paradigm for building trustworthy, high-performance LLM applications. However, its unique, composite architecture (retrieval, reranking, generation) introduces unprecedented challenges for system optimization. The central thesis of this paper is that existing generic LLM traces fail to capture the RAG-specific workload dynamics, creating a significant performance gap between academic research and practical deployment.

To bridge this gap, we design and open-source \sysname, a publicly available RAG workload trace dataset derived from an university-wide Q\&A service that has served over 40,000 students and faculties since April 2024. Through an in-depth analysis of this dataset, we have quantified critical characteristics of real-world RAG workloads: a highly skewed hot document access pattern and significant inter-request temporal locality. These findings provide clear empirical support and research direction for RAG system optimizations, such as content-aware batching and retrieval caching.

\newpage
\bibliographystyle{unsrt}
\bibliography{references}{}

\end{document}